\documentclass[10pt,twocolumn,letterpaper]{article}

\usepackage[pagenumbers]{cvpr}

\definecolor{cvprblue}{rgb}{0.21,0.49,0.74}
\usepackage[pagebackref,breaklinks,colorlinks,allcolors=cvprblue]{hyperref}
\usepackage{multirow}
\usepackage{mathcomp}
\usepackage{amsmath,amssymb}
\usepackage[accsupp]{axessibility}

\title{Profile-Specific 3DMM Regression from a Single Lateral Face Image}

\author{Taiki Kanaya \qquad Hideo Saito\\
Keio University\\
{\tt\small taiki0223@keio.jp}}

\begin{document}
\maketitle
\begin{abstract}
Single-image 3D face reconstruction is a core problem in computer vision, with important clinical applications such as cephalometric landmark analysis in orthodontics. Traditionally, this analysis relies on lateral X-ray imaging; however, frequent X-ray exposure is impractical due to radiation concerns. While recent research has explored detecting landmarks from lateral RGB images as an alternative, existing methods typically rely on 2D features such as the eyes, mouth, ears, and boundary silhouettes, failing to fully exploit the underlying 3D facial geometry spanning the facial profile and jawline, which is essential for accurate diagnosis. Meanwhile, although 3D face reconstruction from frontal views has seen significant progress, most learning-based 3D morphable model (3DMM) regressors are developed and benchmarked on near-frontal images, where appearance cues are abundant. In extreme profile views (yaw $\approx 90^\circ$), much of the face is occluded, and the available signal is dominated by boundary cues, making accurate 3D reconstruction challenging. In this paper, we bridge this gap with geometry-conditioned synthetic data and a simple profile-specific FLAME regression baseline for single lateral images. We introduce ProfileSynth, a dataset created by sampling FLAME shape and pose parameters in extreme yaw ranges and generating photorealistic profile images using a diffusion model conditioned on depth and normal maps. We further study a profile-specific baseline with visibility-aware jawline regularization. Our framework provides a practical baseline for ``profile $\times$ 3DMM'' reconstruction and a promising foundation for more accurate, non-invasive cephalometric analysis from lateral RGB images.
\end{abstract}
    
\section{Introduction}
\label{sec:intro}

The ability to reconstruct 3D facial geometry from profile views is critical for addressing pressing medical challenges in orthodontics. Cephalometric landmark analysis is an indispensable procedure for craniofacial assessment and treatment planning; however, it traditionally relies on lateral X-ray imaging \cite{Lindner2016,King2022}. The need for frequent X-ray examinations poses a significant clinical problem because of the risks of cumulative radiation exposure, especially for younger patients. While recent research has explored detecting landmarks from lateral RGB images as a non-invasive alternative \cite{Takahashi2023,Shimamura2024}, these methods typically rely on 2D appearance features such as the eyes, mouth, ears, and boundary silhouettes. Such approaches fail to incorporate the underlying 3D facial geometry of the profile and jawline, which is essential for achieving the level of diagnostic precision required in clinical practice.

Against this medical background, single-image 3D face reconstruction and 3D morphable model (3DMM) regression have emerged as core tasks in computer vision. While these technologies enable diverse applications such as avatar creation and AR/VR telepresence \cite{MorphaboleModel,MorphableModelSurvey2020}, their potential for radiation-free clinical diagnostics remains an important frontier. Recent methods recover plausible 3D geometry from in-the-wild images, but they are typically developed and benchmarked on near-frontal to moderately rotated views \cite{DECA:SIGGRAPH:2021,EMOCA_CVPR_2022,zielonka22mica}. In contrast, extreme profile images (yaw close to $90^\circ$) remain challenging and comparatively underexplored \cite{Martyniuk_2022_CVPR,wang20243d}. At these views, one side of the face is fully self-occluded, dense correspondences largely disappear, and the available visual signal is dominated by boundary cues \cite{wang20243d}.

In this work, we study single-image FLAME regression for extreme profiles \cite{Li_2017_FLAME}. Given an RGB image $I$ with yaw in the range of $85^\circ$ to $95^\circ$, we regress FLAME shape and head-pose parameters and reconstruct a 3D head mesh. We canonicalize profile direction by horizontal flipping so that yaw is positive, and we fix expression to neutral to isolate ambiguity induced by extreme pose. This setting is difficult because interior facial evidence is heavily reduced: many landmarks and textures are partially or fully invisible, while perceived shape is primarily governed by silhouette and jawline geometry. As a result, training signals used in frontal-biased pipelines become less informative, and errors concentrate in profile-critical regions such as chin projection, jaw angle, and contour consistency \cite{wang20243d}. By addressing these geometric details, we aim to provide a robust foundation for more accurate and radiation-free cephalometric assessment.

A central bottleneck is the lack of supervision and evaluation tailored to the extreme-profile regime \cite{Martyniuk_2022_CVPR,REALY}. In-the-wild datasets rarely provide accurate 3D annotations for strict lateral views, making systematic training and benchmarking difficult \cite{RingNet:CVPR:2019,Martyniuk_2022_CVPR}. In addition, dense correspondence and photometric supervision become unreliable under severe occlusion, and naive vertex losses may over-penalize unobserved geometry \cite{wang20243d}. Progress in this regime therefore requires (i) scalable paired data with reliable geometry labels, (ii) learning objectives aligned with visible profile evidence, and (iii) evaluation protocols that emphasize contour and jawline fidelity.

To address these challenges, we propose a profile-focused study built around three components. First, we introduce ProfileSynth, a synthetic dataset specialized for extreme profiles. We sample FLAME parameters, render geometric cues (depth, normals, silhouettes, and landmarks) under a fixed camera, and synthesize photorealistic RGB images conditioned on rendered depth and normal maps using a ControlNet-style generator \cite{LDM_CVPR_2022,ControlNet_ICCV_2023}. This provides large-scale paired data with exact FLAME ground truth.

Second, we establish a profile-focused evaluation protocol that prioritizes contour fidelity. We do not include silhouette terms during training; instead, we use silhouette-based metrics and profile-oriented errors at test time to assess agreement with visible boundary evidence under extreme yaw.

Third, we study a simple profile-specific FLAME regression baseline with visibility-aware jawline regularization, together with parameter and 3D landmark supervision.

Our contributions are summarized as follows:
\begin{itemize}
    \item ProfileSynth, a synthetic dataset for extreme profile views ($\text{yaw}\in[85^\circ,95^\circ]$), providing paired photorealistic RGB images and exact FLAME ground truth with rendered geometry cues.
    \item A profile-focused evaluation protocol that measures contour and jawline fidelity under extreme yaw and enables systematic benchmarking in the complete lateral setting.
    \item A simple profile-specific FLAME regression baseline with visibility-aware jawline regularization for extreme-profile views.
\end{itemize}

Experiments show that the proposed profile-focused training improves extreme-profile reconstruction quality over representative baselines on both synthetic data and a small real-world profile subset.

\section{Related Work}
\label{sec:relatedworks}

\paragraph{3D Morphable Models and Monocular Reconstruction.}
3D Morphable Models (3DMMs) remain a standard prior for monocular face reconstruction, beginning with the original morphable model of Blanz and Vetter \cite{MorphaboleModel} and followed by improved model spaces such as BFM \cite{bfm09} and LSFM \cite{Booth_2016_CVPR}. FLAME \cite{Li_2017_FLAME} extends this line to an articulated head model with identity, expression, and pose parameters and is widely used in modern learning-based pipelines; broader context is summarized in recent surveys \cite{MorphableModelSurvey2020}. Recent single-image methods commonly regress 3DMM/FLAME parameters directly from RGB input, with RingNet \cite{RingNet:CVPR:2019}, Deep3DFaceRecon \cite{Deng_2019_CVPR_Workshops}, DECA \cite{DECA:SIGGRAPH:2021}, EMOCA \cite{EMOCA_CVPR_2022}, MICA \cite{zielonka22mica}, DAD-3DNet \cite{Martyniuk_2022_CVPR}, and Pixel3DMM \cite{pixel3dmmgiebenhain2025} representing strong baselines for in-the-wild reconstruction. Our work adopts the same FLAME-based formulation, but focuses on the strict lateral regime, where self-occlusion removes much of the interior facial evidence used by frontal-biased pipelines.

\begin{figure*}[t]
  \centering
  \includegraphics[width=0.8\linewidth]{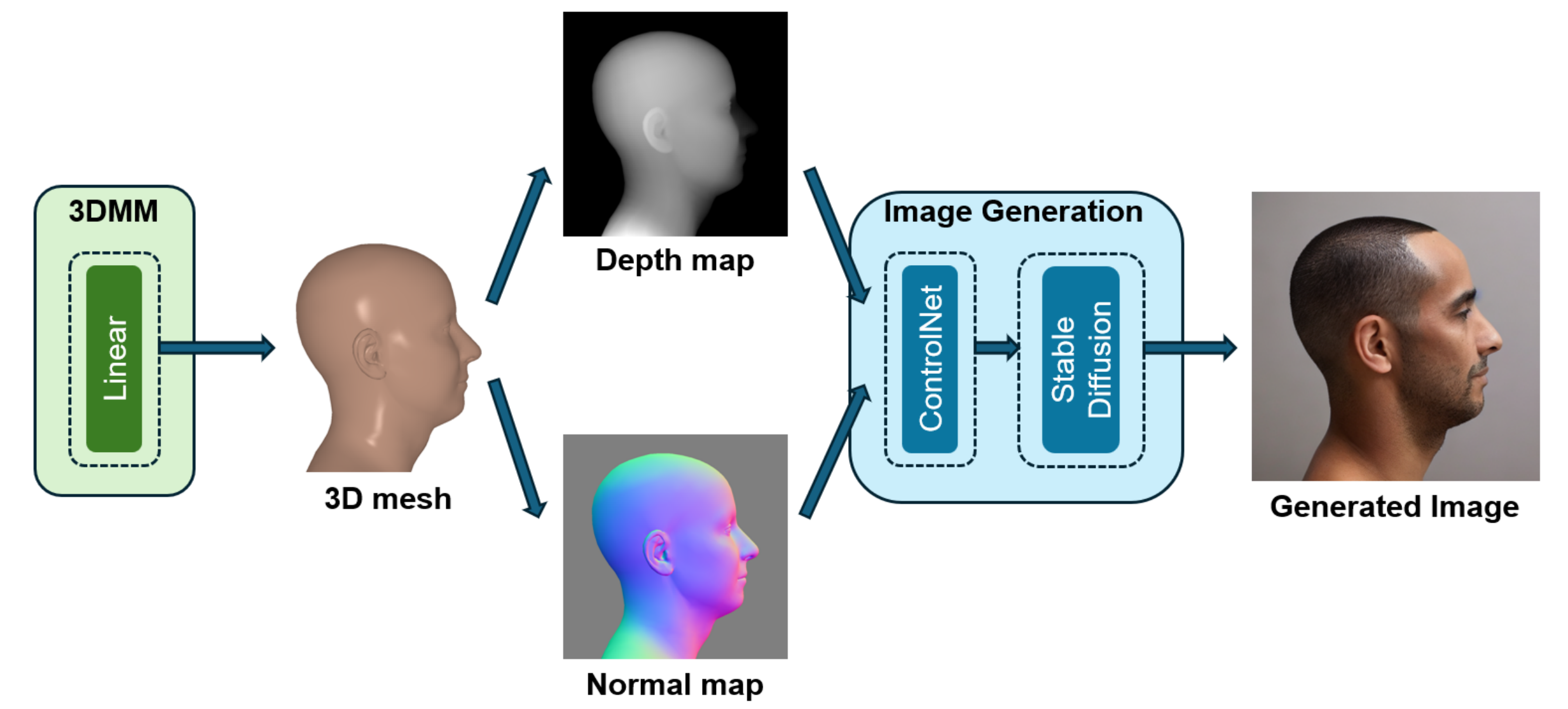}
  \caption{ProfileSynth generation pipeline. We sample FLAME parameters at extreme yaw, render geometry cues (depth, normals, silhouettes, landmarks), and generate photorealistic profile RGB images using ControlNet conditioned on rendered depth and normals.}
  \label{fig:profilesynth}
\end{figure*}

\paragraph{Large-Pose and Profile-Focused Face Reconstruction.}
The difficulty of profile views has long been recognized in large-pose alignment and dense correspondence. 3DDFA \cite{3ddfa}, 3DDFA\_V2 \cite{guo2020towards}, SynergyNet \cite{wu2021synergy}, PRNet \cite{PRNET}, and DAD-3DNet \cite{Martyniuk_2022_CVPR} all address wide pose variation by combining 3D shape priors with pose-aware supervision or dense prediction. These methods demonstrate that profile reconstruction is feasible, but strict near-$90^\circ$ views remain challenging because many landmarks and textures become invisible and contour cues dominate \cite{wang20243d}. Our method is complementary to this line: instead of proposing a new dense correspondence formulation, we specialize a direct FLAME regressor, the training data distribution, and the supervision signals for the complete lateral setting.

\paragraph{Evaluation Under Occlusion and Extreme Yaw.}
Monocular 3D face reconstruction is commonly evaluated on scan-based benchmarks such as NoW \cite{RingNet:CVPR:2019}, while recent analyses such as REALY \cite{REALY} show that reported rankings can depend on alignment choices and the region emphasized by the metric. For profile views, this issue is particularly important because global surface distances may under-emphasize the visible contour, which is the dominant geometric cue. We therefore report standard mesh errors together with contour-oriented measures that better reflect silhouette and jawline fidelity under extreme yaw.

\paragraph{Synthetic Data and Geometry-Conditioned Generation.}
High-quality 3D face datasets such as FaceScape \cite{FaceScape} are valuable for learning geometry, but strict lateral images paired with consistent 3DMM labels are still scarce. This motivates synthetic data generation targeted to the profile regime \cite{Wood_2021_ICCV,synthface}. Latent Diffusion Models \cite{LDM_CVPR_2022} and ControlNet \cite{ControlNet_ICCV_2023} enable image synthesis conditioned on spatial geometry cues, while differentiable rendering toolkits such as PyTorch3D \cite{ravi2020pytorch3d}, nvdiffrast \cite{nvdiffrast2020}, Neural Mesh Renderer \cite{Kato_2018_CVPR}, and Soft Rasterizer \cite{Liu_2019_ICCV} provide practical infrastructure for rendering supervision signals and estimating visibility. We build on this ecosystem to construct ProfileSynth and to train a profile-specific regressor with exact mesh-level labels.

\section{Method}
\label{sec:method}

\subsection{Task Definition}

Given a single RGB image $I$ depicting an extreme-profile face, our goal is to regress FLAME parameters \cite{Li_2017_FLAME} and reconstruct a 3D head mesh. We focus on near-lateral views with yaw in $[85^\circ,95^\circ]$.
Let $\beta$ denote FLAME shape parameters and $\theta$ denote pose parameters. The regressor predicts $(\hat{\beta},\hat{\theta})$ from a preprocessed image $I'$:
\begin{equation}
(\hat{\beta},\hat{\theta}) = g(f(I')),\quad (\hat{V},\hat{L})=\mathrm{FLAME}(\hat{\beta},\hat{\theta}).
\end{equation}
We fix expression to neutral and use a fixed perspective camera for synthetic generation and synthetic evaluation. To remove left-right ambiguity, we canonicalize profile direction by horizontal flipping so that yaw is positive. Extreme profile reconstruction is challenging due to severe self-occlusion and reduced interior appearance cues, making boundary-driven geometry (silhouette and jawline) the dominant source of information.

\subsection{Overview}
Our approach follows a direct FLAME regression pipeline and introduces three profile-focused components: (i) a synthetic dataset (ProfileSynth), (ii) visibility-aware jawline supervision, and (iii) a profile-oriented evaluation protocol.

\subsection{Profile3DMM Regressor}
\begin{figure*}[t]
  \centering
  \includegraphics[width=0.8\linewidth]{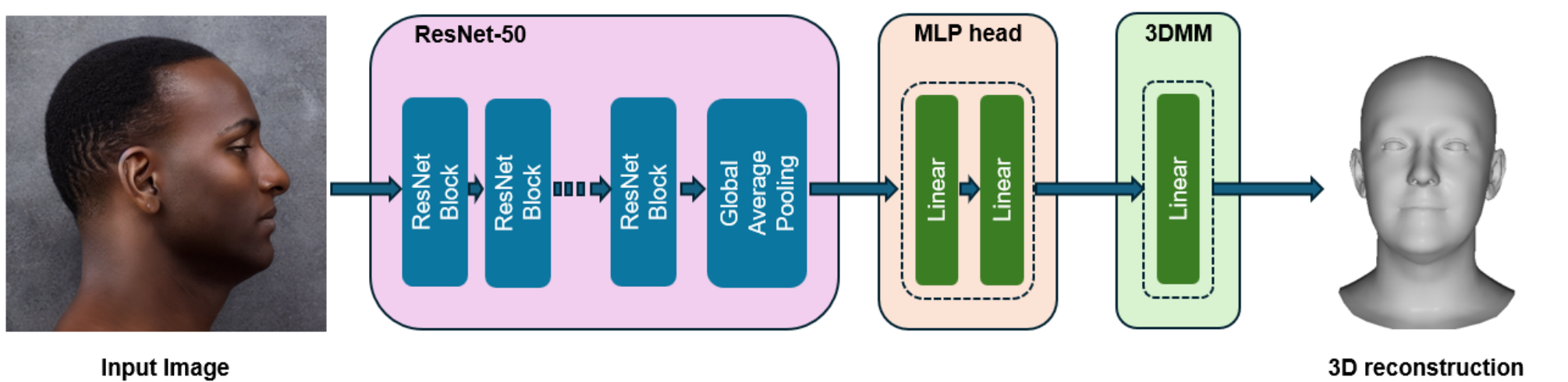}
  \caption{Overview of the proposed Profile3DMM regressor. An ImageNet-pretrained ResNet-50 extracts global features, an MLP head predicts FLAME parameters, and the FLAME decoder outputs the reconstructed mesh and landmarks for supervision and evaluation.}
  \label{fig:method_overview}
\end{figure*}
We use an ImageNet-pretrained ResNet-50 \cite{ResNet_CVPR_2016,ImageNetCVPR09} backbone $f(\cdot)$ to extract image features and an MLP head $g(\cdot)$ to regress FLAME parameters:
\begin{equation}
  z = f(I'), \quad (\hat{\beta},\hat{\theta}) = g(z),
\end{equation}
where $I'$ is the preprocessed input image. The FLAME decoder maps $(\hat{\beta},\hat{\theta})$ to reconstructed vertices $\hat{V}$ and landmarks $\hat{L}$. The MLP head is a two-layer perceptron with a 1024-dimensional hidden layer and predicts 300 shape parameters and 6 pose parameters from the globally pooled ResNet feature, where the pose consists of FLAME global rotation (3) and neck rotation (3), both in axis-angle form. Consistent with our neutral-expression setting, we do not regress expression parameters.

\paragraph{Input Preprocessing}
We resize images to $512 \times 512$, normalize pixel values to $[0,1]$, and apply ImageNet channel normalization. We use face-centered cropping: synthetic images are centered by construction, while real images are manually cropped to keep the full head region near the image center.

\paragraph{Camera Model}
For ProfileSynth, we use a fixed perspective camera with intrinsics identical to those used during data generation. For real-world evaluation on NoW, we follow the official protocol with 7-landmark rigid alignment.

\subsection{ProfileSynth: Profile-Specific Synthetic Dataset}

Inspired by rendering-to-real data generation strategies such as SynthFace \cite{synthface}, we synthesize profile RGB images with ControlNet conditioned on FLAME-rendered geometry. Unlike SynthFace \cite{synthface} and broader synthetic-data face analysis pipelines \cite{Wood_2021_ICCV}, which target general face analysis or image synthesis, we focus on the extreme-profile regime ($\text{yaw}\approx90^\circ$) to support profile-specific 3DMM regression and evaluation.

Learning and evaluation in profile views are difficult with common in-the-wild datasets because strict lateral images with accurate 3D labels are scarce. We therefore construct synthetic training and evaluation data with known geometry labels.

We sample FLAME shape parameters from a clipped Gaussian $\mathcal{N}(0,0.7^2)$ truncated to $[-2,2]$, sample yaw uniformly from $[85 \tcdegree, 95 \tcdegree]$, sample the remaining pose components from the same clipped Gaussian, and fix expression parameters to neutral. For each sample, we render depth maps, normal maps, silhouette masks, and 2D/3D landmarks at $1024\times1024$ using PyTorch3D \cite{ravi2020pytorch3d} with a fixed perspective camera (distance 0.8, FoV $20 \tcdegree$). Conditioned on rendered depth and normals, we generate photorealistic RGB images using a Stable Diffusion v1.5 backbone with ControlNet \cite{LDM_CVPR_2022,ControlNet_ICCV_2023} and use these images as regressor inputs. Each image is paired with the FLAME parameters and rendered geometric signals of the source sample under the same camera model used in training and synthetic evaluation.
Additional generation settings and prompt details are provided in the supplementary material.

\paragraph{Geometry--Appearance Consistency (Sanity Check).}
A potential concern for diffusion-based synthesis is label misalignment: the generated appearance may deviate from the conditioning geometry, leading to noisy supervision.
To sanity-check label consistency in \textit{ProfileSynth}, we measure whether strong image edges in the synthetic RGB remain near the conditioned silhouette boundary.
Specifically, we (i) resize the GT silhouette to the RGB resolution (nearest neighbor), (ii) extract the silhouette boundary, (iii) consider only a narrow band around the boundary to suppress background edges, (iv) extract strong Sobel edges from the generated RGB within the band, and (v) compute boundary-to-edge distances.
As a negative control, we evaluate the same RGB image against a \emph{shifted} silhouette obtained by translating the GT silhouette by a fixed offset (dx=+16 px, dy=0) at the evaluation resolution.
On 10{,}000 test samples, the matched silhouette yields substantially smaller distances and higher near-boundary coverage than the shifted control, suggesting non-trivial geometry--appearance consistency of the synthesized RGB (Table~\ref{tab:rgb_boundary_consistency}).

\begin{table}[t]
  \centering
  \caption{Boundary consistency between synthetic RGB and the conditioning silhouette on the ProfileSynth test split (10{,}000 samples).
  We report the mean boundary-to-edge distance (px; lower is better) and boundary coverage within 2 px (higher is better).
  The shifted control uses a fixed translation of the GT silhouette (dx=+16 px, dy=0) at the evaluation resolution.}
  \label{tab:rgb_boundary_consistency}
  \small
  \setlength{\tabcolsep}{4pt}
  \begin{tabular}{@{}lcc@{}}
    \toprule
    Setting & Mean dist.\,(px) $\downarrow$ & Cov.@2 px (\%) $\uparrow$ \\
    \midrule
    Matched silhouette (ours) & 10.05 & 56.7 \\
    Shifted silhouette (+16 px) & 32.21 & 30.7 \\
    \bottomrule
  \end{tabular}
\end{table}

\subsection{Visibility-Aware Jawline Supervision and Losses}
In profile views, silhouette and jawline geometry dominate visual perception; however, large self-occluded regions make naive vertex supervision unstable.
We therefore emphasize jawline geometry while masking invisible regions.

\paragraph{Visibility}
We rasterize the ground-truth mesh with the ground-truth camera using a $256\times256$ PyTorch3D rasterizer to obtain visibility masks. We treat a jawline vertex as visible if it belongs to at least one visible rasterized triangle, yielding $J_{vis} \subset J$, where $J$ is the set of all jawline vertices. Jawline vertices are defined as FLAME vertices on the jawline region from chin to ear.

\paragraph{Losses}
We use the weighted sum of the following losses:
\begin{equation}
  \mathcal{L} = w_p \mathcal{L}_{param} + w_l \mathcal{L}_{lm3d} + w_j \mathcal{L}_{jaw}
\end{equation}
where $\mathcal{L}_{param}$ is the L2 loss on FLAME parameters, $\mathcal{L}_{lm3d}$ is the L2 loss on 3D landmarks,
and $\mathcal{L}_{jaw}$ is the visibility-aware jawline vertex loss, defined as
\begin{equation*}
\begin{aligned}
  \mathcal{L}_{param} &= \| \hat{\beta} - \beta \|_2^2 + \| \hat{\theta} - \theta \|_2^2, \\
  \mathcal{L}_{lm3d} &= \frac{1}{K} \sum_{k=1}^{K} \| \hat{L}_k - L_k \|_2^2, \\
  \mathcal{L}_{jaw} &= \frac{1}{|J_{vis}|} \sum_{i \in J_{vis}} \| \hat{V}_i - V_i \|_2^2.
\end{aligned}
\end{equation*}
where $\hat{V}_i$ and $V_i$ are predicted and ground-truth vertex positions, respectively, and $K$ is the number of landmarks. We use static 3D landmarks from FLAME for $\mathcal{L}_{lm3d}$ because FLAME dynamic landmark embeddings are designed for near-frontal views and are limited to $|\text{yaw}| \leq 39 \tcdegree$.
In all experiments, we set $(w_p,w_l,w_j)=(1,100,10)$. For $\mathcal{L}_{lm3d}$, we exclude the 17 contour landmarks from the standard 68-point configuration and supervise the remaining 51 static landmarks. We use a fixed jawline set $J$ of 65 vertices.
We do not include silhouette terms in training; silhouette is used only for evaluation.

\section{Experiments}
We target the \emph{extreme-profile} regime (yaw $\approx 90^\circ$), where contour and jawline recovery is critical under severe self-occlusion. Our primary benchmark is ProfileSynth, which provides exact FLAME ground truth and enables systematic contour-oriented evaluation. As a preliminary real-world transfer study, we additionally report results on a small profile subset of the NoW dataset rather than the full benchmark.

\subsection{Datasets}
\paragraph{ProfileSynth.}
ProfileSynth is the dataset used in this paper, generated with ControlNet \cite{ControlNet_ICCV_2023} conditioned on FLAME-rendered depth and normal maps. It contains 100{,}000 samples with yaw uniformly sampled from $[85^\circ,95^\circ]$, shape and non-yaw pose components sampled from a clipped Gaussian distribution, and expression fixed to zero. Input resolution is $512\times512$. We split the dataset into 80{,}000/10{,}000/10{,}000 train/val/test samples.

\paragraph{NoW profile subset.}
For real-world evaluation, we manually select 26 profile images from the NoW dataset \cite{RingNet:CVPR:2019}. We include images if (i) both eyes are not simultaneously visible and (ii) the nose tip lies on the silhouette boundary. Since preprocessing is not fully standardized across methods, we treat these results as indicative only.

\subsection{Baselines}
We compare against representative single-image 3D face reconstruction methods with public implementations.
For quantitative evaluation on ProfileSynth, we compare only methods whose outputs can be represented on the same FLAME2020 topology as the ground truth, since our profile metrics require vertex-wise jawline and silhouette evaluation on a shared mesh/camera setup. Methods based on different 3DMM topologies (3DDFA\_V2 and SynergyNet) are therefore reported only on the NoW profile subset, where the official scan-based protocol supports cross-topology comparison after rigid alignment.
For all baselines, we use the official code and pretrained weights with default preprocessing, and extract the reconstructed mesh for evaluation.
Because ProfileSynth uses neutral expressions, we evaluate geometry in a neutral-expression setting when forming FLAME meshes (e.g., by zeroing expression parameters when applicable) and do not include view-dependent displacement or detail geometry.

\begin{itemize}
  \item DECA \cite{DECA:SIGGRAPH:2021}: FLAME-based regression with pose and detail modeling.
  \item EMOCAv2 \cite{EMOCA_CVPR_2022,DECA:SIGGRAPH:2021,filntisis2022visual}: a DECA-family method emphasizing expression/identity modeling.
  \item 3DDFA\_V2 \cite{guo2020towards,3ddfa_cleardusk}: a fast and stable dense 3D face alignment / 3DMM regression baseline designed for large pose variation.
  \item SynergyNet \cite{wu2021synergy}: jointly reasons about 3DMM parameters and 3D landmarks to improve monocular facial geometry.
  \item Pixel3DMM \cite{pixel3dmmgiebenhain2025}: predicts pixel-aligned cues and fits FLAME.
  \item MICA \cite{zielonka22mica}: identity-centric regression; does not necessarily output explicit pose/camera.
\end{itemize}

\subsection{Metrics and Evaluation Protocol}
\paragraph{Rigid-aligned evaluation on ProfileSynth.}
Several baselines do not explicitly predict pose/camera parameters. To ensure a fair \emph{shape} comparison across such pose-agnostic methods, we evaluate after rigid alignment performed only at evaluation time \cite{Umeyama1991}.
For the methods included in the ProfileSynth quantitative comparison, all meshes share the FLAME2020 topology and are compared in the same FLAME coordinate system.
Given a predicted mesh $\hat{V}\in\mathbb{R}^{N\times 3}$ and ground truth $V\in\mathbb{R}^{N\times 3}$, we estimate a rigid transform $(R,t)\in SE(3)$ by minimizing
\begin{equation}
(R,t) = \arg\min_{R\in SO(3),\,t}\sum_{i\in\mathcal{I}} \|R\hat{V}_i + t - V_i\|_2^2,
\end{equation}
where $\mathcal{I}$ is a predefined vertex set (we use visible vertices unless otherwise stated).
We then apply $(R,t)$ to the prediction and compute all metrics on the aligned mesh.
This rigid alignment is used only for evaluation and is not applied during training or inference.

\paragraph{3D mesh errors (ProfileSynth).}
We report global and jawline-local vertex errors after alignment, and their visible-only variants. Let $J\subset\{1,\dots,N\}$ denote the jawline-band vertex set.
We report (lower is better):
\begin{equation}
\begin{aligned}
E_{\text{all}} &= \frac{1}{N}\sum_{i=1}^{N}\|R\hat{V}_i+t-V_i\|_2, \\
E_{\text{vis}} &= \frac{1}{|\mathcal{I}|}\sum_{i\in\mathcal{I}}\|R\hat{V}_i+t-V_i\|_2, \\
E_{\text{jaw}} &= \frac{1}{|J|}\sum_{i\in J}\|R\hat{V}_i+t-V_i\|_2, \\
E_{\text{jaw,vis}} &= \frac{1}{|J\cap\mathcal{I}|}\sum_{i\in J\cap\mathcal{I}}\|R\hat{V}_i+t-V_i\|_2.
\end{aligned}
\end{equation}
Unless noted, we use the visible-only variants as the primary 3D metrics due to profile self-occlusion.

\paragraph{Silhouette metrics (ProfileSynth).}
We render silhouettes from the \emph{rigid-aligned} meshes using the same renderer settings for both prediction and ground truth.
Silhouette IoU is computed as
\begin{equation}
  \mathrm{IoU}(S,\hat{S}) = \frac{|S \cap \hat{S}|}{|S \cup \hat{S}|},
\end{equation}
where $S$ and $\hat{S}$ are the ground-truth and predicted silhouette masks.
We additionally report a boundary-based metric, boundary Chamfer, computed by extracting silhouette contours (morphological gradient), converting boundary pixels to 2D point sets, and measuring the symmetric Chamfer distance (normalized by image diagonal).

\paragraph{NoW evaluation.}
For NoW, the ground truth is a 3D face scan registered to each image. We follow the official NoW protocol \cite{RingNet:CVPR:2019,REALY}: predictions are aligned to the ground-truth scan using 7-landmark rigid alignment, and we report per-vertex error after alignment (in millimeters).

The official NoW metric averages scan-to-mesh distances over scan vertices after rigid alignment \cite{REALY}. Because this global surface metric does not explicitly emphasize profile contour fidelity, we complement NoW with contour-oriented metrics on ProfileSynth, where exact silhouettes are available.

\subsection{Implementation Details}
We implement our model in PyTorch and train for 30 epochs with AdamW \cite{LoshchilovH19}, batch size 64, and AMP enabled. The initial learning rates are $1\times10^{-5}$ for the backbone and $1\times10^{-4}$ for the MLP head, with weight decay $1\times10^{-4}$. We select the checkpoint with the lowest validation loss. We set the training seed to 42 for all experiments. Training is performed on a single NVIDIA RTX 4090 GPU using PyTorch/torchvision \cite{torchvision2016}, PyTorch3D \cite{ravi2020pytorch3d}, and diffusers \cite{von-platen-etal-2022-diffusers}. Data generation uses fixed per-sample seeds to ensure reproducibility.

\subsection{Main Results}
\subsubsection{Results on ProfileSynth}
\begin{table}[t]
  \centering
  \caption{Rigid-aligned comparison on ProfileSynth (test, $N=10{,}000$). Lower is better for errors and boundary Chamfer; higher is better for IoU. Vertex errors are measured in FLAME coordinate units. Best results are in bold.}
  \label{tab:profilesynth_results}
  \small
  \setlength{\tabcolsep}{3pt}
  \begin{tabular}{@{}lcccc@{}}
    \toprule
    Method & $E_{\text{vis}}\downarrow$ & $E_{\text{jaw,vis}}\downarrow$ & IoU$\uparrow$ & B-Ch.\ $\downarrow$ \\
    \midrule
    DECA & 0.004225 & 0.007189 & 0.945245 & 0.007064 \\
    EMOCAv2 & 0.004384 & 0.007544 & 0.940235 & 0.007880 \\
    MICA & 0.004344 & 0.005153 & 0.935497 & 0.008757 \\
    Pixel3DMM & 0.007344 & 0.008876 & 0.901414 & 0.013370 \\
    Ours & \textbf{0.001672} & \textbf{0.001758} & \textbf{0.986498} & \textbf{0.001186} \\
    \bottomrule
  \end{tabular}
\end{table}
\begin{table}[t]
  \centering
  \caption{NoW profile subset results (mm, lower is better). Evaluated on $n=26$ images; Pixel3DMM reports on $n=19$ due to failures. Best results are in bold.}
  \label{tab:now_results}
  \small
  \setlength{\tabcolsep}{5pt}
  \begin{tabular}{lcc}
    \toprule
    Method & Median $\downarrow$ & Mean $\downarrow$ \\
    \midrule
    DECA & 1.6147 & 2.0081 \\
    EMOCAv2 & 1.7138 & 2.1651 \\
    MICA & \textbf{0.9432} & \textbf{1.1606} \\
    3DDFA\_V2 & 1.4363 & 2.0531 \\
    SynergyNet & 1.3449 & 1.6747 \\
    Pixel3DMM & 1.4768 & 2.3471 \\
    Ours (Profile3DMM) & 1.3688 & 1.7060 \\
    \bottomrule
  \end{tabular}
\end{table}
Table~\ref{tab:profilesynth_results} reports common-frame shape results on the ProfileSynth test split ($N=10{,}000$) under shared post-hoc rigid alignment.
Profile3DMM achieves the best performance across all metrics, including both visible-only 3D errors and contour-based measures.
Compared to the strongest baseline (DECA), Profile3DMM reduces $E_{\text{vis}}$ by $2.53\times$ and $E_{\text{jaw,vis}}$ by $4.09\times$, improves $\mathrm{IoU}$ by $+0.041$, and reduces boundary Chamfer by $5.95\times$.

\begin{table*}[t]
  \centering
  \caption{Ablation on ProfileSynth (test, no rigid alignment). We report pre-alignment visible-only 3D mesh errors ($E_{\text{vis}}$, $E_{\text{jaw,vis}}$) due to strong self-occlusion in extreme profile views. Lower is better for errors and boundary Chamfer; higher is better for IoU.}
  \label{tab:ablation_profilesynth}
  \small
  \setlength{\tabcolsep}{5.0pt}
  \begin{tabular}{lccccc}
    \toprule
    Variant &
    3D LM L2$\downarrow$ &
    $E_{\text{vis}}\downarrow$ &
    $E_{\text{jaw,vis}}\downarrow$ &
    IoU$\uparrow$ &
    B-Chamfer$\downarrow$ \\
    \midrule
    Baseline (param + 3D lm + jaw) &
    \textbf{0.003782} &
    \textbf{0.003558} &
    \textbf{0.003420} &
    0.979422 &
    0.002114 \\
    w/o 3D landmark loss &
    0.004019 &
    0.003653 &
    0.003649 &
    0.979562 &
    0.002096 \\
    w/o jawline loss &
    0.003786 &
    \textbf{0.003558} &
    0.003436 &
    0.979400 &
    0.002112 \\
    Param-only supervision &
    0.003908 &
    0.003591 &
    0.003562 &
    \textbf{0.979639} &
    \textbf{0.002083} \\
    High jawline weight ($\times 30$) &
    0.003835 &
    0.003583 &
    0.003484 &
    0.979490 &
    0.002099 \\
    \bottomrule
  \end{tabular}
\end{table*}

\subsubsection{Results on NoW profile subset}
Table~\ref{tab:now_results} shows results on the NoW profile subset ($n=26$), following the official NoW evaluation protocol (7-landmark rigid alignment; mm).
These results indicate preliminary but mixed real-domain transfer: our method improves over DECA, EMOCAv2, 3DDFA\_V2, and Pixel3DMM on this limited subset, while MICA and SynergyNet obtain lower overall errors.
Pixel3DMM fails on 7 of the 26 images due to detection failures under extreme profiles, so we report its metrics on the remaining images. Given the small sample size and preprocessing differences across methods, we treat these results as indicative rather than conclusive. They also suggest that facial-outline improvements may be under-reflected by the global scan-based metric, motivating our profile-oriented evaluation on ProfileSynth; the supplementary material provides additional clinical contour-proxy analysis and protocol-stability checks aligned with the paper's medical motivation.
Supplementary clinical contour-proxy evaluation on 2{,}277 validated lateral photographs further supports real-domain transfer, showing lower jawline-contour error than DECA and EMOCAv2.

\subsection{Ablation Study}
\label{subsec:ablation}

We ablate the supervision terms used for training on ProfileSynth in the \emph{extreme-profile} regime (yaw $\approx 90^\circ$). All variants share the same architecture, data split, and optimization settings; only the loss configuration is changed.
Unlike Table~\ref{tab:profilesynth_results}, we report pre-alignment errors because all ablation variants predict in the same canonical FLAME frame and no cross-method pose normalization is needed. We focus on 3D landmark error, visible-only mesh errors, and silhouette-based measures.

\paragraph{3D landmark supervision is a key geometric constraint.}
Removing the 3D landmark loss consistently degrades geometric accuracy.
Compared to the baseline, the 3D landmark error increases (3D LM L2: $0.003782 \rightarrow 0.004019$, $\approx\!+6.3\%$), and both visible-only mesh errors worsen ($E_{\text{vis}}$: $0.003558 \rightarrow 0.003653$, $\approx\!+2.7\%$; $E_{\text{jaw,vis}}$: $0.003420 \rightarrow 0.003649$, $\approx\!+6.7\%$).
This indicates that sparse 3D landmark supervision provides an effective geometric anchor under severe self-occlusions in profile views.

\paragraph{Jawline supervision exhibits subtle and non-monotonic effects.}
Removing the jawline term leads to only marginal changes relative to the baseline (e.g., $E_{\text{jaw,vis}}$: $0.003420 \rightarrow 0.003436$), suggesting partial redundancy with parameter and landmark supervision in this setting.
Moreover, increasing the jawline weight by $30\times$ does not yield a monotonic improvement on the jawline-band error ($E_{\text{jaw,vis}}$: $0.003484$).
While silhouette metrics change slightly across variants, these results suggest that simply up-weighting a \emph{3D} jawline-vertex objective is not sufficient to consistently improve profile reconstruction, motivating contour-aware objectives that more directly couple jawline geometry to the visible silhouette boundary.

\paragraph{Param-only supervision is insufficient for 3D geometry.}
Using only parameter supervision degrades visible 3D accuracy ($E_{\text{vis}}$: $0.003591$; $E_{\text{jaw,vis}}$: $0.003562$), even though it slightly improves silhouette metrics (IoU and boundary Chamfer).
This highlights a potential metric trade-off in extreme profiles: contour alignment can improve without a corresponding gain in 3D geometric fidelity, reinforcing the importance of explicit geometric supervision for reliable profile reconstruction.

\subsection{Qualitative Results}
\begin{figure*}[t]
  \centering
  \includegraphics[width=0.8\linewidth]{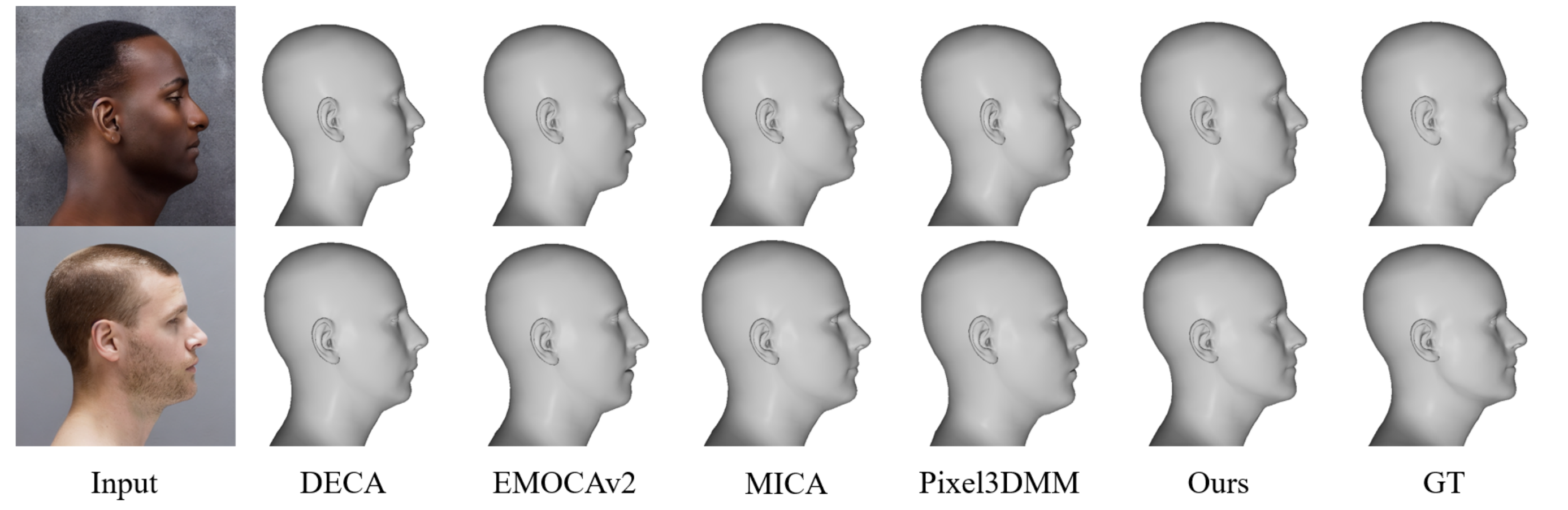}
  \caption{Qualitative comparison on ProfileSynth. We show reconstructions from single extreme-profile images (yaw $\approx 90^\circ$). All predicted meshes are rigidly aligned to the ground-truth (GT) mesh and rendered from the same canonical profile view. Our method better preserves the facial outline and jawline and produces shapes closer to GT than existing baselines.}
  \label{fig:qualitative}
\end{figure*}

\begin{figure*}[t]
  \centering
  \includegraphics[width=0.8\linewidth]{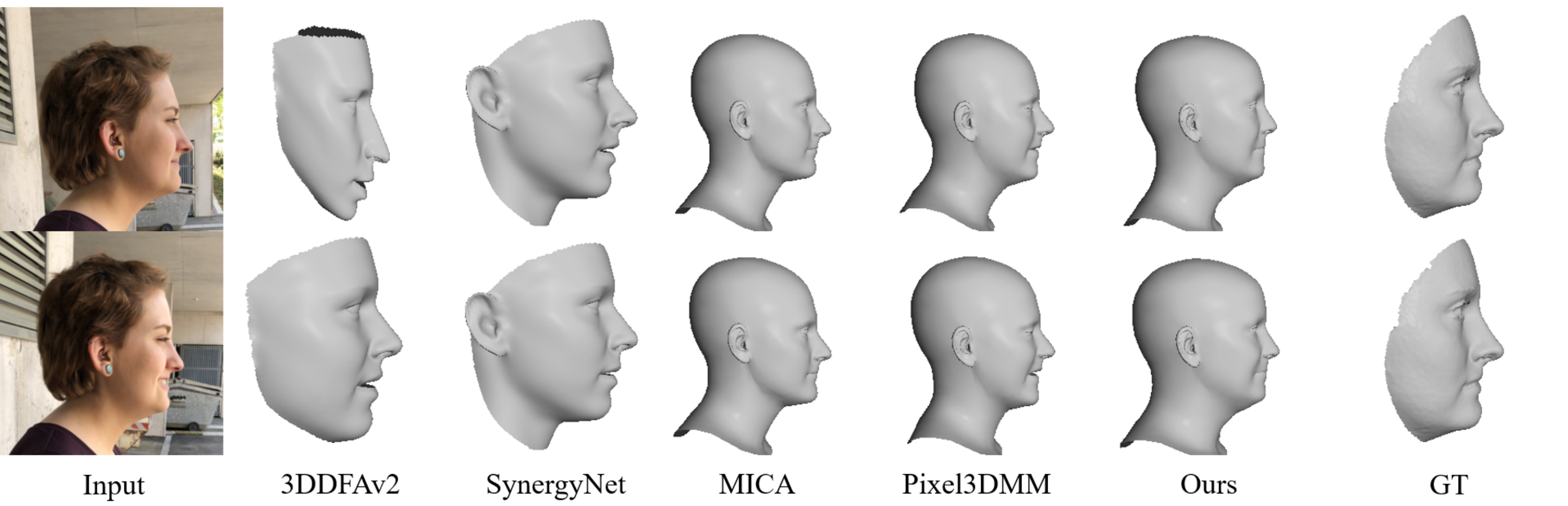}
  \caption{
Qualitative comparison on the NoW dataset~\cite{RingNet:CVPR:2019}.
We show two images of the same subject: \emph{neutral} (top) and \emph{expressive} (bottom).
All baselines are run with their official implementations and default settings.
NoW provides a single subject-level 3D scan as ground truth (GT) even for expressive inputs, so expression-aware methods may deviate around the mouth.
Visually, our method yields the cleanest and most consistent chin-to-neck profile contour.
}
\label{fig:now_qual}
\end{figure*}

\paragraph{Qualitative comparison on ProfileSynth}
Figure~\ref{fig:qualitative} visualizes representative reconstructions on ProfileSynth under extreme-profile conditions (yaw $\approx 90^\circ$). We rigidly align each prediction to the ground truth and render all methods from the same canonical profile camera to emphasize shape quality. Compared with DECA~\cite{DECA:SIGGRAPH:2021}, EMOCAv2~\cite{EMOCA_CVPR_2022}, MICA~\cite{zielonka22mica}, and Pixel3DMM~\cite{pixel3dmmgiebenhain2025}, our method more faithfully recovers the facial outline and jawline.

\paragraph{Qualitative comparison on NoW.}
Figure~\ref{fig:now_qual} compares our profile-specific reconstruction with 3DDFA\_V2, SynergyNet, MICA, and Pixel3DMM on the NoW dataset.
We follow a standard evaluation protocol and run all baselines with their official code and default settings (i.e., without manually setting expression parameters to zero).
Because NoW provides a single subject-level 3D scan as GT even for expressive images \cite{RingNet:CVPR:2019}, reconstructions that preserve input-dependent facial expressions can exhibit deformations around the mouth region that deviate from the GT scan.
Despite this challenge, our method produces more stable head-shape reconstructions across neutral and expressive inputs, with a cleaner chin-to-neck profile contour than competing methods.

\section{Discussion and Conclusion}
\label{sec:conclusion}

\subsection{Limitations}

The present study is limited to neutral expression, a fixed camera model, and a modest real-world evaluation set. Although ProfileSynth enables controlled training and benchmarking, a synthetic-to-real gap likely remains, especially because of profile-specific failures in detection, cropping, and appearance variation. A larger real-world benchmark that explicitly targets strict profile reconstruction would make the evaluation more conclusive.

\subsection{Conclusion}

We studied single-image FLAME regression in the extreme-profile regime (yaw $\in [85^\circ,95^\circ]$), where severe self-occlusion weakens interior appearance cues and makes contour information especially important. To make this setting easier to study systematically, we introduced a profile-focused pipeline consisting of ProfileSynth, a synthetic dataset with paired FLAME labels, a profile-focused evaluation protocol, and a simple baseline with visibility-aware jawline regularization. Because competing methods use heterogeneous pose/camera parameterizations, our main quantitative comparison is deliberately shape-focused under shared rigid alignment.

Experiments on ProfileSynth show that this setup yields substantially better reconstruction quality than representative baselines under rigidly aligned evaluation, especially on visible-region and silhouette-based measures. Our real-image results on a small NoW profile subset are preliminary and mixed, but they suggest at least partial transfer beyond the synthetic domain \cite{RingNet:CVPR:2019}; supplementary contour-proxy results on 2{,}277 validated clinical photographs further support this trend. Overall, this work provides a focused baseline for studying 3D face reconstruction in the complete lateral setting.

\paragraph{Data and code availability.}
The clinical photographs used in the supplementary evaluation were collected under institutional ethics approval and cannot be redistributed.
Because ProfileSynth is derived from FLAME~\cite{Li_2017_FLAME}, redistribution of the full synthetic dataset is subject to the FLAME license.
We will release the generation pipeline, training code, and pretrained weights; the dataset can be fully reproduced by users who obtain the FLAME model from the official distribution (\url{https://flame.is.tue.mpg.de}).
We also plan to release synthetic assets generated using publicly available FLAME resources where licensing permits.

\section*{Acknowledgment}
The authors thank C.~Tachiki and Y.~Nishii (Department of Orthodontics, Tokyo Dental College) for providing the clinical profile photographs used in the supplementary evaluation.

{
    \small
    \bibliographystyle{ieeenat_fullname}
    \bibliography{mybib}
}

\clearpage
\appendix
\maketitlesupplementary

\section{ProfileSynth Construction}
\label{sec:suppl_profilesynth}

The main paper introduces \textit{ProfileSynth} as the source of paired supervision for extreme-profile FLAME regression. This supplement records the concrete generation settings relevant to reproducibility. Table~\ref{tab:suppl_profilesynth_config} summarizes the fixed data-generation configuration used to create the synthetic profile regime studied in the paper.

\begin{table}[t]
    \centering
    \caption{ProfileSynth generation setup.}
    \label{tab:suppl_profilesynth_config}
    \small
    \setlength{\tabcolsep}{4pt}
    \begin{tabular}{@{}lp{0.56\linewidth}@{}}
        \toprule
        Item & Setting \\
        \midrule
        Face model & FLAME2020 \cite{Li_2017_FLAME} \\
        Requested sample count & 100{,}000 \\
        Shape dimensionality & 300 \\
        Expression & fixed to zero \\
        Yaw range & $[85^\circ,95^\circ]$ \\
        Other pose components & clipped Gaussian sampling \\
        Camera & fixed perspective, distance $0.8$, FoV $20^\circ$ \\
        Render resolution & $1024 \times 1024$ \\
        Input resolution & $512 \times 512$ \\
        Conditioning signals & depth + normal \\
        Diffusion backbone & Stable Diffusion v1.5 \cite{LDM_CVPR_2022} \\
        ControlNet branches & depth + normal \cite{ControlNet_ICCV_2023} \\
        ControlNet model IDs & \begin{tabular}[t]{@{}l@{}}\texttt{lllyasviel/}\\ \texttt{control\_v11f1p\_sd15\_}\\ \texttt{depth}\\ \texttt{lllyasviel/}\\ \texttt{control\_v11p\_sd15\_}\\ \texttt{normalbae}\end{tabular} \\
        Diffusion settings & 25 steps, guidance $7.5$, cond. scales $0.7/0.4$ \\
        Saved assets & RGB, silhouette, depth, normal, 2D/3D landmarks, FLAME parameters, and camera parameters \\
        \bottomrule
    \end{tabular}
\end{table}

Each sample is generated by first sampling FLAME shape and pose parameters in the extreme-profile regime, rendering depth maps, normal maps, silhouettes, and landmarks under a fixed camera, and then synthesizing RGB appearance with ControlNet conditioned on the rendered depth and normal maps. The prompt family is intentionally narrow: it requests photorealistic side-profile portraits with neutral expression, straight-ahead gaze, visible ear and jawline, tied-back hair, no glasses or mask, no hands near the face, and soft even lighting on a plain background, while the negative prompt suppresses low-resolution, non-photographic styles, frontal or three-quarter views, strong occlusions, harsh lighting, and obvious facial distortions. This narrow prompting is deliberate: the paper does not aim to model the full diversity of in-the-wild portrait photography, but rather to create a controlled profile-only supervision source.

\section{Geometry--Appearance Consistency}
\label{sec:suppl_consistency}

Because the training RGB images are diffusion-generated, a key concern is label misalignment between the conditioning geometry and the synthesized appearance. To test this, we measure whether strong edges in the generated RGB remain near the conditioning silhouette boundary. As in the main paper, we compare the matched silhouette against a shifted-control silhouette obtained by translating the ground-truth mask.

\begin{table}[t]
    \centering
    \caption{Geometry--appearance consistency on the ProfileSynth test split ($10{,}000$ samples). Lower is better for distances; higher is better for coverage.}
    \label{tab:suppl_boundary_consistency}
    \small
    \setlength{\tabcolsep}{3pt}
    \begin{tabular}{@{}lccc@{}}
        \toprule
        Setting & Mean & Cov.@2px & Sym. Chamfer \\
        \midrule
        Render vs.\ GT & 0.565 & -- & -- \\
        Synth vs.\ matched & 10.047 & 0.567 & 5.722 \\
        Synth vs.\ shifted & 32.214 & 0.307 & 17.250 \\
        \bottomrule
      \end{tabular}
    \end{table}
    
    \begin{table*}[t]
        \centering
        \caption{Ablation study on ProfileSynth (test split, no rigid alignment). Lower is better for errors and boundary Chamfer; higher is better for IoU.}
        \label{tab:suppl_ablation}
        \small
        \setlength{\tabcolsep}{5pt}
        \begin{tabular}{lccccc}
            \toprule
            Variant & 3D LM L2$\downarrow$ & $E_{\text{vis}}\downarrow$ & $E_{\text{jaw,vis}}\downarrow$ & IoU$\uparrow$ & B-Chamfer$\downarrow$ \\
            \midrule
            Baseline (param + 3D lm + jaw) & \textbf{0.003782} & \textbf{0.003558} & \textbf{0.003420} & 0.979422 & 0.002114 \\
            w/o 3D landmark loss & 0.004019 & 0.003653 & 0.003649 & 0.979562 & 0.002096 \\
            w/o jawline loss & 0.003786 & \textbf{0.003558} & 0.003436 & 0.979400 & 0.002112 \\
            Param-only supervision & 0.003908 & 0.003591 & 0.003562 & \textbf{0.979639} & \textbf{0.002083} \\
            High jawline weight ($\times 30$) & 0.003835 & 0.003583 & 0.003484 & 0.979490 & 0.002099 \\
            \bottomrule
        \end{tabular}
    \end{table*}
    The matched setting is substantially better than the shifted control on all reported measures. This does not imply that the synthetic RGB is perfectly realistic; rather, it supports the narrower and more important claim that the generated profile appearance remains sufficiently tied to the conditioning geometry to serve as supervised training data.
    
    \section{Profile-Aware Supervision}
    \label{sec:suppl_loss}
    
    The training loss used in the paper is
    \begin{equation}
      \mathcal{L} = w_p \mathcal{L}_{param} + w_l \mathcal{L}_{lm3d} + w_j \mathcal{L}_{jaw},
    \end{equation}
    with $(w_p,w_l,w_j)=(1,100,10)$. The parameter term supervises FLAME shape and pose, the landmark term supervises 51 static non-contour landmarks, and the jawline term supervises a fixed FLAME jawline-band vertex set (65 vertices in our implementation).
    
Two design choices are important in the extreme-profile setting. First, we exclude the 17 contour landmarks from the standard 68-point configuration because contour definitions become unstable near complete profile views and FLAME dynamic landmarks are not designed for this regime. Second, the jawline loss is applied only to the \emph{visible} subset of jawline-band vertices, where visibility is determined by rasterizing the ground-truth mesh under the ground-truth camera. This prevents the supervision from over-penalizing self-occluded geometry that is not directly supported by the input image.

The main paper already reports this compact ablation; here we emphasize a cautious interpretation. Table~\ref{tab:suppl_ablation} shows that 3D landmark supervision contributes most clearly: removing it degrades landmark and visible-region geometry, indicating that sparse 3D anchors remain important even when contour cues dominate. Param-only supervision is also worse than the baseline on landmark and jawline-band errors, even though it slightly improves the silhouette metrics. By contrast, the standalone effect of the jawline term is modest in this summary: removing it changes the reported metrics only slightly, and increasing its weight does not improve the jawline-band error monotonically. We therefore view the jawline term as a visibility-aware profile regularizer rather than as the dominant source of the gain. This is consistent with the central message of the paper: profile reconstruction benefits from profile-specific supervision, yet contour-aware reconstruction is not solved by simply up-weighting a local 3D vertex loss.

\section{Evaluation Scope and Real-Image Transfer}
\label{sec:suppl_eval}

The synthetic benchmark remains the primary evidence of the paper because it provides exact FLAME ground truth and exact silhouettes under strict profile views. For cross-method comparison on ProfileSynth, we evaluate after rigid alignment to isolate shape quality from differences in predicted pose or camera parameterization across methods. We then report both visible-region 3D errors and silhouette-based measures, since contour fidelity is the dominant observable cue under near-$90^\circ$ yaw.

The NoW profile-subset experiment serves a different purpose. It is intended only as a preliminary transfer study showing whether a profile-specific synthetic prior can generalize beyond the synthetic domain. Because the subset is small and the official NoW metric is a global scan-based error after alignment, it should not be interpreted as a contour-specific benchmark. For this reason, the supplementary clinical proxy evaluation below is useful: it is still not a substitute for exact 3D ground truth, but it probes the profile contour directly rather than through a global scan distance.

\section{Supplementary Clinical Contour-Proxy Evaluation}
\label{sec:suppl_clinical}

Because the paper is motivated by radiation-free cephalometric analysis from lateral facial photographs, we additionally conducted a supplementary evaluation on real clinical profile photographs from the same institutional image collection used in recent cephalometric photograph studies \cite{Takahashi2023,Shimamura2024}. The images were made available to us for research use under the same data-governance framework as those studies. This experiment is not used as the main benchmark of the paper, but it is informative as a real-domain check of whether the learned profile prior improves clinically relevant visible contour reconstruction.

Since these clinical photographs do not provide mesh-level 3D ground truth, we evaluate with a silhouette-proxy metric focused on the anterior jawline contour. Predicted FLAME silhouettes are aligned to the clinical mask using isotropic scale and translation estimated from an anatomy-focused upper-profile bounding box, and the primary metric is computed on the front-side jawline ROI rather than on the full silhouette. We keep the older full-ROI contour metric only as a diagnostic because it is noticeably more sensitive to posterior-neck and torso contamination.

For mask extraction, we first obtain a person mask and then apply a profile-specific trimming heuristic that suppresses lower-neck and shoulder leakage; GrabCut \cite{Rother2004GrabCut} is used as a fallback when the initial mask is insufficient. Images are excluded when the mask is invalid, the predicted mesh is missing or malformed, the aligned render fails, the jawline ROI is too small to evaluate, or the image is obviously top-cropped. In the validated run, 2{,}362 images are discovered, 2{,}277 are valid, and 85 are skipped (76 non-profile or otherwise unusable images, 7 invalid bounding boxes, and 2 missing or malformed predictions).

The primary metric on the valid subset has mean 60.20 px, median 57.72 px, and standard deviation 27.55 px. A bootstrap analysis gives a 95\% confidence interval of [59.06, 61.35] for the mean and [56.19, 59.73] for the median, indicating that the central tendency of the proxy metric is stable under resampling. We summarize the protocol-stability checks in Table~\ref{tab:suppl_clinical_stability}.

\begin{table}[t]
    \centering
    \caption{Clinical protocol stability checks on the supplementary contour-proxy evaluation. Alignment sensitivity is measured on the 512-sample subset with ROI start fixed at 0.55. Lower is better.}
    \label{tab:suppl_clinical_stability}
    \small
    \setlength{\tabcolsep}{4pt}
    \begin{tabular}{lcc}
        \toprule
        Check & Setting & Mean \\
        \midrule
        Alignment & none & 120.05 \\
        Alignment & bbox & \textbf{59.79} \\
        Alignment & bbox+pca & 125.00 \\
        Sanity & predicted shape & \textbf{59.79} \\
        Sanity & mean shape & 61.29 \\
        Sanity & random shape & 62.40 \\
        \bottomrule
    \end{tabular}
\end{table}

Table~\ref{tab:suppl_clinical_stability} shows that the chosen alignment protocol is not arbitrary: using the upper-profile bounding box reduces the contour error by roughly half relative to no alignment, while adding PCA-based orientation is harmful in this dataset. The same sampled subset also provides a sanity check for the learned prior: the predicted-shape contour is better than both a mean-shape baseline and random-shape samples (paired Wilcoxon $p=1.38\times10^{-9}$ and $1.05\times10^{-14}$, respectively). The diagnostic full-ROI metric is at least 20 px worse than the front-side ROI metric on 14.10\% of valid samples and at least 40 px worse on 2.33\%, which supports the decision to focus the evaluation on the clinically relevant visible jawline.

\begin{table}[t]
    \centering
    \caption{Supplementary clinical contour-proxy comparison on the common valid subset ($n=2{,}277$). Lower is better.}
    \label{tab:suppl_clinical}
    \small
    \setlength{\tabcolsep}{5pt}
    \begin{tabular}{lccc}
        \toprule
        Method & Mean $\downarrow$ & Median $\downarrow$ & Std. $\downarrow$ \\
        \midrule
        DECA & 83.03 & 82.93 & 31.33 \\
        EMOCAv2 & 82.46 & 82.57 & 31.26 \\
        Ours & \textbf{60.20} & \textbf{57.72} & \textbf{27.55} \\
        \bottomrule
    \end{tabular}
\end{table}

Table~\ref{tab:suppl_clinical} shows that our method improves substantially over DECA and EMOCAv2 on this contour-focused proxy metric. On the common validated subset, the paired mean improvement is 22.83 px over DECA and 22.26 px over EMOCAv2; our method is better on 79.49\% and 78.92\% of images, respectively, with paired Wilcoxon $p=3.83\times10^{-164}$ and $6.64\times10^{-161}$. We emphasize that this does not replace exact 3D evaluation: the metric is a carefully designed proxy on visible profile structure. However, it is useful supplementary evidence that the profile-specific prior learned from \textit{ProfileSynth} transfers to real clinical photographs in a way that is consistent with the medical motivation of the paper.

\end{document}